\documentclass[a4paper]{article}
\usepackage{INTERSPEECH2022}
\usepackage[linesnumbered,ruled,vlined,english,onelanguage]{algorithm2e}
\usepackage{graphicx}
\usepackage{grffile} 
\title{Compute Cost Amortized Transformer for Streaming ASR}
\name{Yi Xie, Jonathan Macoskey, Martin Radfar, Feng-Ju Chang, Brian King, Ariya Rastrow,\\ Athanasios Mouchtaris, Grant P. Strimel}
\address{
 Alexa Machine Learning, Amazon, USA }
\email{\{yixiey, macoskey, radfarmr, fengjc, bbking, arastrow, mouchta, gsstrime\}@amazon.com}

\begin{document}

\maketitle
\begin{abstract}
We present a streaming, Transformer-based end-to-end automatic speech recognition (ASR) architecture which achieves efficient neural inference through compute cost amortization. 
Our architecture creates sparse computation pathways dynamically at inference time, resulting in selective use of compute resources throughout decoding, enabling significant reductions in compute with minimal impact on accuracy.
The fully differentiable architecture is trained end-to-end with an accompanying lightweight arbitrator mechanism operating at the frame-level to make dynamic decisions on each input while a tunable loss function is used to regularize the overall level of compute against predictive performance.
We report empirical results from experiments using the compute amortized Transformer-Transducer (T-T) model conducted on LibriSpeech data.
Our best model can achieve a $60\%$ compute cost reduction with only a $3\%$ relative word error rate (WER) increase.
\end{abstract}

\section{Introduction} 
The rapidly expanding adoption of voice assistant systems has coincided with trending approaches for speech and natural language processing (NLP) implemented with end-to-end, fully neural architectures~\cite{graves2013speech,rao2017exploring,kannan2019large}. However, real-time execution and compute resource considerations have challenged the speech community to adopt these neural approaches to realize their predictive accuracy while addressing the high compute requirements for neural network inference~\cite{li2019improving}.

Meanwhile, the marked success of Transformers in NLP~\cite{vaswani2017attention,devlin2018bert} has motivated researchers to modify and apply these architectures for ASR systems~\cite{zhang2020transformer,yeh2019transformer,vila2018end,nakatani2019improving,gulati2020conformer,huang2020conv}. Given the superior performance of capturing temporal contextual information from streaming input, the application of Transformer-based ASR has been demonstrated to achieve lower WER compared to recurrent neural network (RNN) architectures~\cite{zhang2020transformer,huang2020conv,gulati2020conformer,karita2019comparative}. Nevertheless, the additional compute demand of the Transformer, such as the quadratic computation complexity of the core self-attention mechanism, introduces further obstacles of model scalability and can hinder production deployments.
Therefore, it is desirable to design efficient inference approaches for Transformer-based ASR which enable the models to perform in common virtual assistant settings, such as executing on resource constrained, low-power edge devices in real-time.

 Since its inception, various approaches have been proposed for reducing Transformer cost where most focus on scaling the self-attention mechanism. For example, Performer~\cite{choromanski2020rethinking} and Linformer~\cite{wang2020linformer} leverage low-rank approximations of the self-attention matrix to lighten both memory and computation cost. Another prominent technique is to naturally sparsify the attention matrix by limiting the field of view to fixed and predefined patterns such as local sliding windows~\cite{zaheer2020big} and blockwise paradigms~\cite{qiu2019blockwise}. Meanwhile,~\cite{tay2020sparse,kitaev2020reformer} exploit forms of learnable sparsity patterns by sorting/clustering the input values. The direction of these approaches is still to apply fixed patterns with consideration of the global view of the sequence. We refer readers to~\cite{tay2020efficient} for a thorough survey and summary of efficient Transformer methods. While delivering impressive results, applied in isolation, these methods have limitations for real-time, low-footprint, speech applications which our approach seeks to address.
 \textit{Adaptability}: Deployed, production voice-assistants are required to generalize to a wide spectrum of input varieties. Dimensions such as audio length (short commands or longer dialogues and dictations), quality of audio, and speaker characteristics all differ significantly for each input.
 One-size-fits-all, static approaches can face obstacles in adapting to the various input demands while maximizing model performance.
 %; failing to reduce on-demand computational redundancy conditioned on varying complexity of different input.
 \textit{Locality}: Previous methods emphasize simplifying the scaled dot-product attention operation whereas other components (e.g. feed-forward (FF) layers) also consume a significant proportion of compute cost, especially for shorter inputs. Where prior works focus on reducing attention head cost in isolation, we will emphasize addressing the compute budget by holistically considering the entire network and its composite components in unison.
 \textit{Expressivity}: Ultimately parameter reductions and compression methods for more efficient inference can eventually inhibit modeling capacity in the network, leading to a degradation in predictive performance. We look to reduce compute inference while retaining full model capacity and without any alterations of the underlying Transformer architecture, the approach should be compatible with other compression techniques.

To address these motivations, in this work, we present a Transformer architecture which is able to \textit{amortize compute cost on demand} at inference by dynamically activating compute components conditioned on input frames. 
Our work is inspired by prior investigations of dynamic compute for speech processing which have yielded notable results~\cite{macoskey2021bifocal,shi2021dynamic,weninger2021dual,macoskey2021amortized}.
%sukhbaatar2019adaptive is not a speech paper
Considering speech examples can be lengthy streaming sequences with many frames of different levels of inherent complexity (e.g. consider silent frames between acoustically rich segments), dynamic computing enables a desired balance between accuracy and efficiency by altering the compute expenditure each frame. Unlike existing methods~\cite{macoskey2021bifocal,weninger2021dual,macoskey2021amortized} which bifurcate ASR into only two fixed branched encoder networks, with our approach, an exponential family of dynamic branches are generated from a single Transformer encoder which significantly expands the modeling space and boosts the adaptability and expressivity of the model.
Lightweight prediction arbitrators (e.g. a simple FF or RNN) are devised to holistically amortize compute cost across the whole of the input within the entire Transformer encoder by sketching sparse, yet robust computation pathways on a frame-by-frame basis. Our resulting models therefore achieve significant reductions in compute cost while preserving predictive performance and imposing minimal additional overhead.

%The additional cost brought by the arbitrators are minor compared with the overall compute overhead reduction. 

%The rest of this paper is structured as follows. Transformer basics are discussed in Section~\ref{sec:transformer}. Section~\ref{sec:methodology} introduces the detailed design and optimization strategy of our amortization Transformer. Next, Section~\ref{sec:exp} demonstrates experimental setup and results.

% \end{itemize}

\begin{figure}
\centering
	\includegraphics[width=8cm]{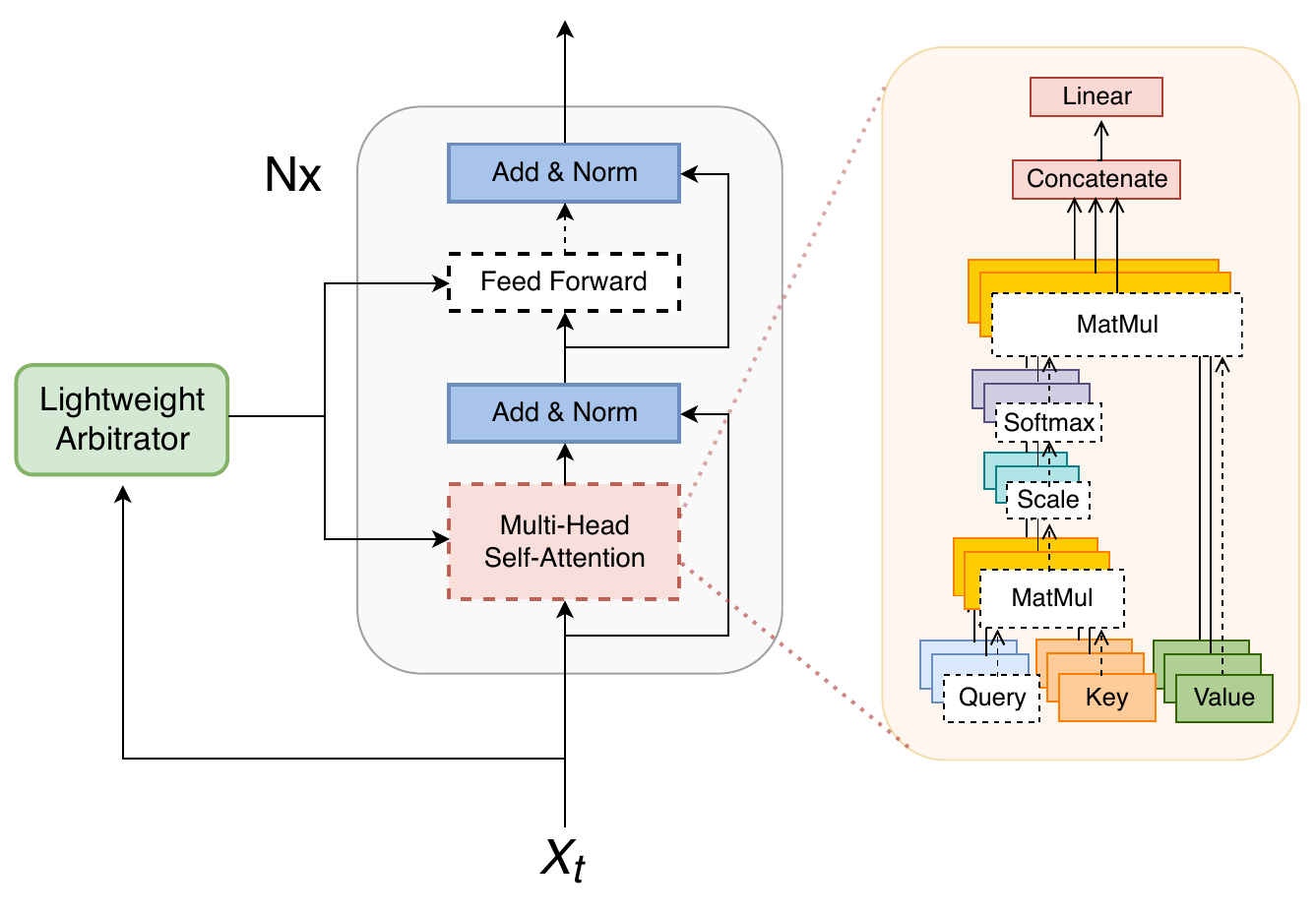}
	\vspace{-4.0mm}
	\caption{Illustration of the arbitrator and Transformer backbone in each block. The lightweight arbitrator toggles whether to evaluate subcomponents during the forward pass.}
	\label{fig:overall} %% label for first subfigure
	\vspace{-4mm}
\end{figure}
\vspace{-2mm}
\section{Transformer Basics}
\label{sec:transformer}
The Transformer architecture is a multi-layered network defined by the Transformer block which is stacked multiple times in succession. The workhorse of the Transformer block consists of its multi-headed self-attention (MHA) layer. The MHA operation is subsequently followed by a FF module with intermittent layer normalization and residual connections. For input $X$ and head attention $h$, three matrices are computed in an MHA layer: the query $Q_h$, the key $K_h$, and the value $V_h$ by
\begin{equation}
{Q_h} = {W_q}X, \quad K_h = {W_k}X, \quad {V_h} = {W_v}X
\end{equation}
from learned weight matrices $W_q$, $W_k$, and $W_v$ of dimension $( \frac{d}{H} \times d)$, where $d$ is the representation dimension and $H$ is the number of attention
heads. The queries and keys are pairwise compared and scored across input pairs. The computed values are then
weighted through a Softmax operation on the attention logits $P_h$ to derive the attention output $A_h$:

\begin{equation}
A_h = \text{Softmax}(P_h)V_h= \text{Softmax}(\alpha{{Q_h}{K_h}^T})V_h.
\label{eq:softmax}
\end{equation}
The $\alpha$ is a scaling factor typically set to $\frac{1}{\sqrt{d}}$. The output of the attention heads $A_1, \dots, A_H$ are then concatenated together and passed into a linear projection $Y = W_o\left[A_1 \cdots A_H\right]$.
$Y$ is then layer normalized and a skip connection added to arrive at the attention output $X_A$. It is then fed position-wise through a $d_{f}$-dimension $\text{FF}_1$ layer followed by a $d$-dimension $\text{FF}_2$ layer as $\text{FF}_2(\text{FF}_1(X_A))$ before a final layer normalization and skip connection.

\vspace{-2mm}
\section{Compute Cost Amortized Transformer}
\label{sec:methodology}
%\subsection{Overall Structure}

The fundamental component of our proposed framework is shown in Figure~\ref{fig:overall} which depicts the $i$-th Transformer block and a lightweight arbitrator network $D(\cdot)$. The arbitrator is the core of the amortization mechanism, and it is much lighter in size compared to the Transformer encoder such that it does not introduce significant compute overhead at inference time. The arbitrator takes an input $x^t$, the audio feature frame at timestep $t$, and predicts a set of toggle decisions.
Toggles turn-off components of the transformer block by effectively zeroing-out their contributions to the forward pass.
Toggling operates on both the FF module (FF$_1$ + FF$_2$) and the MHA component of the Transformer block to create sparse compute pathways.

In our design, we consider each FF module as an integrated unit. Namely, when toggling-off is triggered for the FF module on a particular frame, the FF module evaluation would be skipped over, or rather, the output to it would be identical to its input because of the residual shortcut.
As a result, each frame in which a FF module is toggled off, two $O(d \cdot d_{f})$ floating point operations (FLOPs) are eliminated from inference.

Next, we will introduce the details of our toggle approaches for the MHA module.

\begin{figure*}[t]
	\centering
	%\subfigure[Experimental setup]{
		\label{key}
		\includegraphics[width=0.82\linewidth]{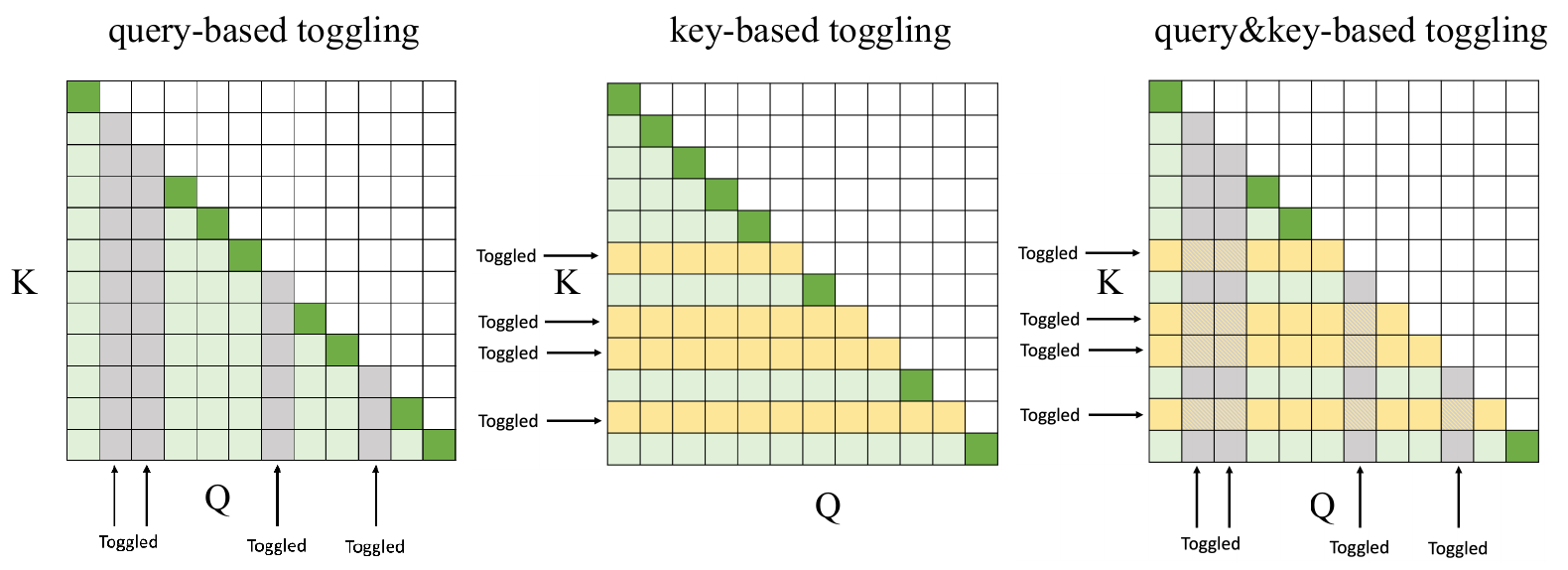}
		\vspace{-2mm}
	\caption{Illustration of query-based toggling, key-based toggling, and query\&key-based toggling on attention maps.}
	\vspace{-3mm}
	\label{fig:toggling}
\end{figure*}

\subsection{Multi-headed Attention Toggling}

The bulk of the neural inference cost for Transformer MHA layers can be attributed to three sources. The first source is the $W_q$, $W_k$ and $W_v$ weight matrix operations performed on each input across all attention
heads and layers.
The second and more commonly discussed source comes from the quadratic scaling (in input
length) across attention heads where pairwise interactions are computed with the dot product between all inputs for each layer and attention head. 
The last source is the final linear projection $W_o$. In our design, to make the calculation in MHA modules more efficient and well-structured, we introduce two types of toggling to control these computations: query-based and key-based.

In query-based toggling, the arbitrator can decide to ignore the output from an attention head at timestep $t$ by simply replacing the result with a zero-vector.
As a result, $W_q$ need not be applied to the input and a component of the projection $W_o$ for the head can be skipped as well, saving two $O(d^2/H)$ operations.
In a complementary fashion, with key-based toggling, the arbitrator can decide to ignore keys in the attention mechanism.
As a result, neither the key computed with $W_k$ nor the values with $W_v$ need be computed at that timestep, saving again two $O(d^2/H)$ operations.
Both toggles also save computation on the scaled attention as well.
Query-based toggling achieves it immediately, by not computing attention on the current frame, while key-based toggling eliminates an extra dot product computation for each future frame.
Thus, the toggling will inject the sparsity into the attention matrix, where each non-contributing cell saves $O(d/H)$ operations.
See Figure~\ref{fig:toggling}, which shows the visualization of the proposed toggle strategies on the attention matrix.
Interestingly, the impacts in terms of compute reductions have varying dynamics based on a toggle's type and positional occurrence.
A query toggle off saves more compute near the end of the sequence, as it gets to ignore all past history, while the key-toggle pays the most dividends if done early in the sequence, where each subsequent frame gets to benefit from the savings.
Our arbitrators learn to apply both key-based and query-based toggling in conjunction to capitalize on the benefits of both.

%On the other hand, in key-based toggling $(W_k \times X_t)$ and $(W_v \times X_t)$ are not computed. Attention is still computed at this input, but other frames' attention will not attend to $X_t$. This method will inject sparsity into the attention matrix and reduce the memory footprint, not requiring the historical storage of $K_h$ of timestep $t$. For query/key-based toggling, we combine query-based toggling and key-based toggling together, and trigger the two types of toggling individually. By applying our proposed toggling strategies, computation cost from all three primary sources are addressed, leading to more efficient self-attention.

%Fig~\ref{fig:toggling} shows the visualization of our proposed toggle strategies. Query-based toggling can be applied on those frames which contain less essential information. Skipping the attention output for these frames will save a notable percentage of calculations without significantly affecting the final ASR output. In the case of applying key-based toggling, the specific frame performs more independently which has less correlation to other frames. And finally, query/key toggling combines the two for the greatest reduction in computation cost.

\subsection{End-to-End Optimization}

In our amortization framework, we aim to design the arbitrator to be jointly optimized with the Transformer network.
The arbitrator network $D(\cdot) = \left(D_f(\cdot), D_q(\cdot),  D_k(\cdot)\right)$ is decomposable into three functions representing each of the corresponding toggle types.
Hence, the arbitrator produces the decisions (probabilities) at each timestep $t$ for retaining or discarding compute components for FF modules, queries, and keys as $p_f^t=D_f(x^{t})\in \left[0,1\right]^{B}$, $p_q^t=D_q(x^{t})\in \left[0,1\right]^{B\times H}$ and $p_k^t=D_k(x^{t})\in \left[0,1\right]^{B\times H}$ where $B$ and $H$ denote the number of Transformer blocks and the number of heads, respectively. 
Since the decisions emitted from the arbitrator network are not binary (with ideally $p=0$ indicating skip and $p=1$ for compute), they are passed to a sampler,
which is responsible for converting those decisions to harder decisions $s^t$.
During training time, we apply the Gumbel-Softmax/Sigmoid~\cite{jang2016categorical} technique to generate differentiable samples from the arbitrator's soft probability outputs.

% \begin{equation}
% \textbf{M}_{f}^{t} = \sigma(\frac{\log(p) +{\varepsilon_1}-{\varepsilon_2}}{\tau})
% \end{equation}

% where $\varepsilon$ is sampled from Gumbel-Softmax distribution~\cite{gumbel1960bivariate}, $s$ is a factor to control the amount of the added Gumble noise and the temperature factor ${\tau}$ controls the discreteness of the sampling. Specifically, as $\tau$ approaches 0, samples from the Gumble-Softmax distribution becomes identical to categorical distribution and the decision mask becomes one-hot. During training, we make $\tau$ gradually close to 0 to coerce the produced masks towards more discrete decisions.
The arbitrator output is used to create simple masks for each component during training which are used to efficiently train the full architecture end-to-end. For example, an attention mask $M_h$ for head $h$ of an attention block would be defined with entries $(t,j)$ as
\begin{equation}
  M_h^{(t,j)}=\begin{cases}
              -\infty \quad\quad\text{if $t < j$, enforcing causality}\\
               \ln(s^t_k)\quad\text{otherwise}
            \end{cases}
\end{equation}
where $s^t_k$ is the sampled key toggle decision at time $t$. We can then update Equation~\ref{eq:softmax} with simply\footnote{In practice we found masking both key and value computations added robustness during training.}

\begin{equation}
A_h = \text{Softmax}(P_h + M_h)V_h.
\end{equation}
With the application of masking, the forward pass of our model is completed with little additional training overhead for computing the arbitrator results and mask generation. The entire training procedure is differenitiable and both the arbitrator and main Transformer network learned jointly.

To control the total amount of compute used at inference, we regularize the decisions of the arbitrator with a modified objective function:
\begin{equation}
\mathcal{L} = \mathcal{L}_{transducer} + \beta \cdot \mathcal{L}_{compute},
\end{equation}
which includes the standard neural transducer loss~\cite{macoskey2021bifocal} and an added compute cost penalty term $\mathcal{L}_{compute}$. 
Accounted for by the cumulative number of FLOPs across the components of the network for a streaming sequence,
$\mathcal{L}_{compute}$ drives more computation cost reduction while maintaining predictive performance of the model. Including the scaling-factor $\beta$ enables us to balance this trade-off and even anneal this value during training for greater stability and superior model convergence. See Section~\ref{sec:exp} for experimental details.

\vspace{-2mm}
\section{Experimental Results}
\label{sec:exp}

\subsection{Experimental Setup}

\begin{table}
\centering
\caption{Transformer encoder parameter setup.}
\vspace{-2mm}
\begin{tabular}{cc}
\hline
%\multicolumn{2}{c}{Transformer encoder parameter setup} \\ \hline
Input feature/embedding size           & 516            \\ 
MLP size                               & 1024           \\
\# layers                           & 12             \\
\# heads per layer                  & 4              \\
Output size                            & 512            \\ \hline
\end{tabular}
\vspace{-2mm}
\label{tab:model}
\end{table}

\begin{table}
\centering
\caption{Memory and computation overhead.}
\vspace{-2mm}
\begin{tabular}{ccc}
\hline
Model               & Params & FLOPs/frame \\ \hline
Transformer Encoder & 35M    & 34M   \\

Single-FF Arb.         & 0.08M  & 0.08M \\
Dual-FF Arb.            & 0.16M  & 0.16M \\ 
Single-RNN Arb.          & 0.5M   & 0.5M  \\
Dual-RNN Arb.            & 1M     & 1M    \\\hline
\end{tabular}
\label{tab:cost}
\vspace{-2mm}
\end{table}

\begin{table*}[t!]
\centering
\caption{The results of the amortized streaming T-T on the LibriSpeech dataset with different toggle strategies and arbitrator structures and baseline models. CCR Ratio indicates compute cost reduction ratio compared with the original dense Transformer encoder.}
\vspace{-2.0mm}
\begin{tabular}{ccccccc}
\hline
           & \multicolumn{2}{c}{Query-based} & \multicolumn{2}{c}{Key-based} & \multicolumn{2}{c}{Query\&key-based} \\ \hline
Arbitrator & WER          & CCR Ratio        & WER         & CCR Ratio       & WER             & CCR Ratio          \\ \hline
Single-FFN Arb. & 8.51\%        & 30.61\%          & 8.44\%       & 33.45\%         & 8.59\%           & 41.96\%           \\
Dual-FFN Arb.   & 8.60\%        & 29.62\%          & 8.21\%       & 28.83\%         & 8.57\%           & 55.52\%           \\
Single-RNN Arb. & 8.64\%        & 37.18\%          & 8.74\%       & 38.16\%         & 8.74\%           & 52.37\%             \\
Dual-RNN Arb.   & 8.54\%        & 34.80\%          & 8.58\%       & 31.44\%         & \textbf{8.39\%}  & \textbf{60.31\%}  \\ \hline
Full Causal  & \multicolumn{3}{c}{WER: 8.12\% } & \multicolumn{3}{c}{CCR Ratio: N/A} \\
Sliding Windows  & \multicolumn{3}{c}{WER: 8.82\% } & \multicolumn{3}{c}{CCR Ratio: 12.93\%} \\
Random Toggling  & \multicolumn{3}{c}{WER: $>$ 20\% } & \multicolumn{3}{c}{CCR Ratio: $\sim$ 60\%} \\ \hline

\end{tabular}
\label{tab:result}
\vspace{-4mm}
\end{table*}

\textbf{Dataset:} We investigate our model using the LibriSpeech corpus~\cite{panayotov2015librispeech} which contains 960 hours of training data and the evaluation is performed on a ``clean'' test data set. To preprocess the audio clips, a 64-dimensional log-filterbank energy feature extractor is applied on all data. These feature frames are stacked with a stride size of two and downsampled by three to produce 30ms frames.\\
\\
\noindent \textbf{Model Setup:} Our baseline model is a streamable Transformer-Transducer (T-T)~\cite{zhang2020transformer} architecture with a Transformer as the encoder and a two-layer 1024-unit LSTM prediction network. The detailed configuration for the Transformer encoder is listed in Table~\ref{tab:model}. 
We implement two types of arbitrator designs: a single and a dual implementation.
In the single arbitrator design, one arbitrator network is used to enact the amortized toggling for the Transformer encoder given the initial audio features. 
With the dual implementation, we break the encoder stack into bottom blocks and top blocks.
We use one arbitrator to operate on the bottom blocks from the input signal and use the second arbitrator on the top blocks with ingesting as input the output from the lower blocks.
All arbitrators are implemented with either a two-layer, 128-unit FF module or a two layer, 128-unit LSTM module. 
The arbitrator networks consist of an output projection layer to the proper toggle dimension to produce amortization decision masks.
As we can see from Table~\ref{tab:cost}, the arbitrators (Arb.) are considerably light-weight compared to our main Transformer encoder.\\

%Additionally, considering multiple blocks are concatenated in the encoder, we further design a dual-arbitrator structure to summary the computation plans. In this case, the Transformer encoder is evenly chucked into bottom section and top section with equal number of blocks. A bottom arbitrator is responsible for bottom blocks and the top arbitrator predicts the decisions for the top section. In particular, the input of bottom arbitrator is same as the input of the Transformer encoder and the feature representation of the intermediate layers at the intersection of bottom and top blocks is sent to the top arbitrator as input. We list the additional memory and computation cost of utilizing arbitrators in our proposed framework. As we can see from Table~\ref{tab:cost}, the arbitrators are very light-weighted compared with our Transformer backbone even in the dual-arbitrator setting, especially considering the computational overhead saved by eliminating the redundant computation in Transformer encoder.

\noindent \textbf{Training Details:} To train the amortized T-T, we divide the learning procedure into two primary phases: 1) \textit{pre-training} --  the T-T model and arbitrators are jointly optimized by minimizing RNN-T loss for first 120 epochs with $\beta$ set to 0, enforcing no compute cost penalization during this phase, and 2) \textit{fine-tuning} -- after pre-training, we fine-tune the model to converge with the compute cost penalty, and the Gumbel sampling is activated. We initially set $\beta=1\mathrm{e}{-8}$ and gradually increase it to $5\mathrm{e}{-8}$, while the Gumbel sampling is enabled and used for the final 80 epochs of training with its temperature linearly annealed from $1$ to $1\mathrm{e}{-5}$ and its contribution scaling factor annealed from $0$ to $1$. By annealing the Gumbel sampling parameters, we switch the arbitrator output transitions from ``softer'' to ``harder'' decisions along training. 

For all trainings, the Adam optimizer~\cite{kingma2014adam} is applied and the learning rate schedule is ramped up linearly during first 16K steps. We use a vocabulary of 4097 word pieces and the standard RNN-T beam search decoding with a width of 16 to obtain WER measurements as our accuracy metric. All the models for experiments presented in this paper are trained on $3\times8$ NVIDIA Tesla V100 GPUs with a per-core batch size of 16 (effective batch size of 384).

\vspace{-2mm}
\subsection{Main Results}
We train the full causal T-T model and report a WER of 8.12\% as our main baseline. 
We also build baseline models using a causal T-T with a limited view via the sliding windows where the attention is computed only on the current frame and 10 previous frames~\cite{zhang2020transformer}, and a baseline which relies on purely randomized toggling to achieve its compute reduction.

For our compute amortized T-T, the WER and the compute cost reduction of Transformer encoder for different model settings are reported in Table~\ref{tab:result}.
We varied the experiments using different MHA toggle settings to use only query-based, only key-based and both query\&key-based.
The sampling parameters are set to use hard toggle decisions for all evaluations such that true sparse pathways are created by discarding compute operations in the forward pass.
Our best model using query\&key-based toggling with a dual-RNN arbitrator can achieve a significant $60.31\%$ compute cost reduction while only incurring a minor $3\%$ increase in relative WER compared to the baseline model.
Furthermore, our approach significantly outperforms the sliding window baseline which only reduces $12.93\%$ compute cost while causing $8.6\%$ relative WER increase.

%The computation densities of MHA and FF after the pre-training phase and fine-tuning phase over an utterance example.
\begin{figure}
\vspace{-2.5mm}
	\centering
		\includegraphics[width=0.85\linewidth]{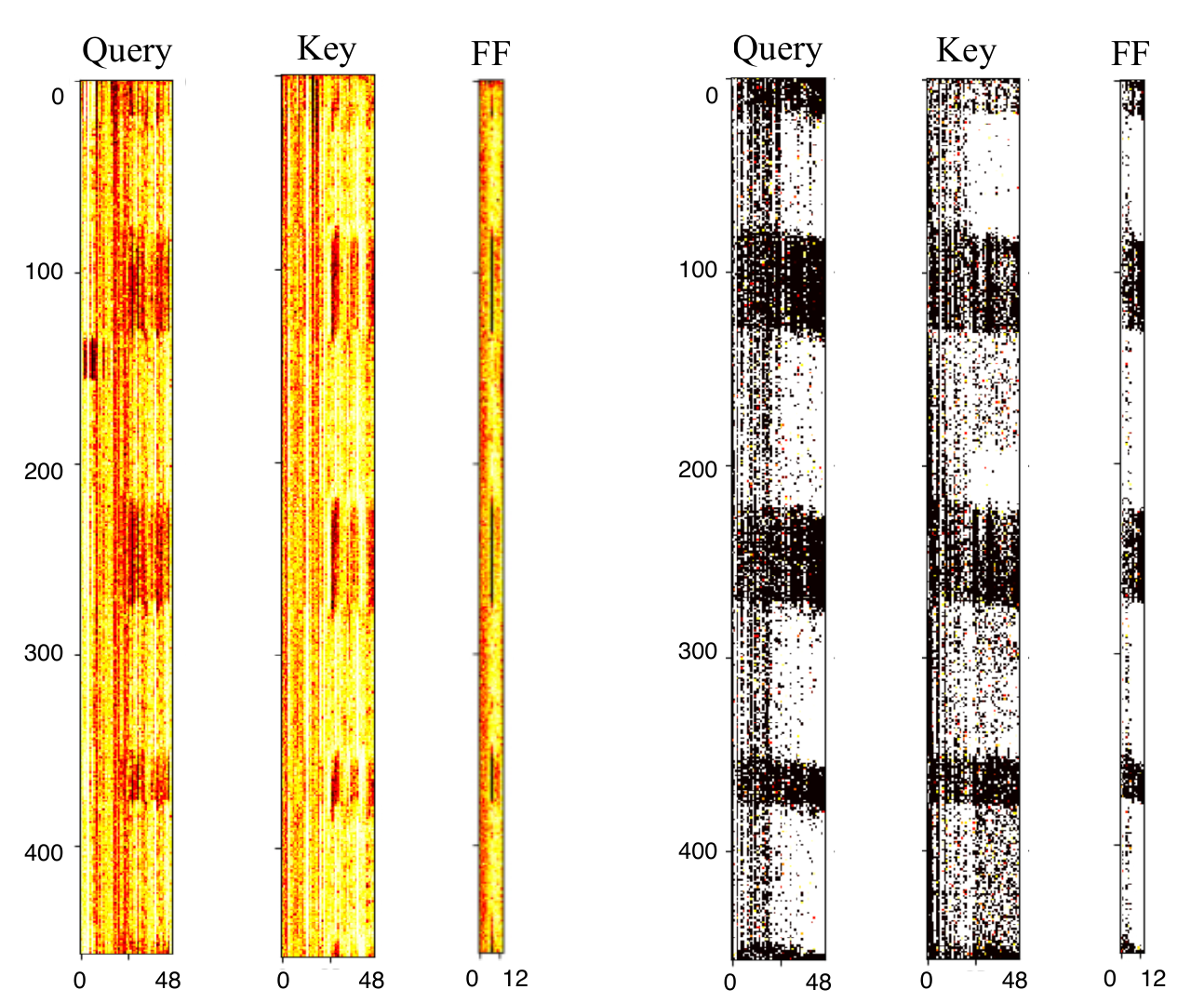}
	\vspace{-2.5mm}
	\caption{The computation densities of MHA and FF after pre-training and fine-tuning phases over an utterance example. The y-axis reflects the audio frames. The x-axis of heatmaps represents 48 attention heads and 12 FF modules in the encoder respectively. Darker areas associate with less compute cost. }
	\label{fig:heat}
	\vspace{-6mm}
\end{figure}

To further investigate the behavior of our proposed compute cost amortized Transformer, we visualize in Figure~\ref{fig:heat} the arbitrator output from an example utterance. 
The vertical axes of these heatmaps represent the frames of the audio streaming while the horizontal axes represent the network layers/MHA heads from lowest to highest levels.
We include both soft decision and hard decision heatmaps where the darker regions signal toggle-off decisions. We can observe clear amortization patterns emerging in many dimensions. 
First, we find that the arbitrator can learn to toggle off for lower complexity frames, particularly silence regions between more acoustically rich areas. 
Second, we find that query-based toggling off has a higher occurrence than key-based on the lower levels near the signal representation. Conversely, key-based toggling off has a higher occurrence than query-based in the higher levels near the output of the network.
Intuitively this aligns with the idea that there is compressability/redundancy with higher level representations in the top layers of the network which therefore can be stored sparsely in the accumulated history via key toggling off frames.
Additionally, one observes overall more toggling off occurring in the lower blocks generally for MHA where typically, the bottom representations are closer to the raw features than the top layer representations. 
Last, we see more key computations are eliminated at the end of the utterance.
This also aligns with intuition because for late keys, there are not many future frames in which their keys would be required for attention.

\vspace{-1mm}
\section{Conclusion}
\vspace{-1mm}
In this work, we propose to improve the compute cost for Transformer-based streaming ASR models using amortization mechanisms. For each input frame, a sparse computation pathway is created in a dynamic fashion based on the binary decision output from lightweight prediction arbitrators. Extensive experimental results demonstrate that our models are able to yield significant compute cost reductions with negligible prediction performance loss. 
\clearpage

\bibliographystyle{IEEEtran}

\bibliography{main}

\end{document}